\documentclass[11pt,letterpaper]{article}

\usepackage[margin=1in]{geometry}
\usepackage{times}
\usepackage{latexsym}
\usepackage[T1]{fontenc}
\usepackage[utf8]{inputenc}
\usepackage{microtype}

\usepackage{amsmath}
\usepackage{amssymb}

\usepackage{graphicx}
\usepackage{booktabs}
\usepackage{multirow}
\usepackage{xcolor}
\usepackage{caption}

\usepackage{url}
\usepackage[colorlinks=true,
            citecolor={blue!60!black},
            linkcolor={blue!60!black},
            urlcolor={blue!60!black}]{hyperref}

\usepackage[authoryear,round,sort&compress]{natbib}
\newcommand{\pbs}{\textsc{PBS}}
\newcommand{\ccp}{\textsc{CCP}}
\newcommand{\effsw}{\textsc{Eff-Sw}}
\newcommand{\Rone}{R1}

\title{More Thinking, More Bias:\\
Length-Driven Position Bias in Reasoning Models}

\author{%
  Xiao Wang  \\ \texttt{FujianAI42@163.com}
}

\date{}

\begin{document}
\maketitle
\begin{abstract}
Chain-of-thought (CoT) reasoning and reasoning-tuned models such as
DeepSeek-\Rone{} are commonly assumed to reduce shallow, heuristic
biases by thinking carefully.
We test this on \emph{position bias} in multiple-choice QA and find a
different story: within any reasoning-capable model, per-question
position bias scales with the length of the reasoning trajectory.
Across thirteen reasoning-mode configurations
(two \Rone-distilled 7--8B models, two base models prompted with CoT,
and DeepSeek-\Rone{} at 671B) on MMLU, ARC-Challenge, and GPQA, twelve
show a positive partial correlation
$\rho(\text{length}, \pbs \mid \text{accuracy})$, ranging from 0.11 to
0.41 (all $p < 0.05$).
All twelve open-weight reasoning-mode configurations show
monotonically increasing \pbs{} across length quartiles; a truncation
intervention provides causal evidence that continuations from later
points in the trajectory are increasingly likely to shift toward
position-preferred options (16\% $\rightarrow$ 32\% for
\Rone-Qwen-7B).
At 671B, aggregate \pbs{} collapses to 0.019, but the length effect
still manifests in the longest quartile (\pbs{}~=~0.071), suggesting
accuracy gates the \emph{expression} of length-driven bias rather
than eliminating the underlying mechanism.
We additionally find that \emph{direct-answer} position bias is a
distinct phenomenon with a different footprint (strong in
Llama-Instruct-direct, weak in Qwen-Instruct-direct, and uncorrelated
with trajectory length): CoT reasoning replaces this
\emph{baseline bias} with \emph{length-accumulated bias}.
Our results argue that reasoning-capable models should not be treated
as order-robust by default in MCQ evaluation pipelines, and offer a
diagnostic toolkit (\pbs{}, \ccp{}, effective switching, truncation
probes) for auditing position bias in reasoning models.
\end{abstract}

\section{Introduction}
\label{sec:intro}

Reasoning-tuned language models~--- OpenAI's o-series,
DeepSeek-\Rone{} \citep{deepseek-ai-2025-r1}, Qwen's QwQ family
\citep{qwen-team-2024-qwq}, and their distilled derivatives~--- are routinely
promoted as models that ``think longer to answer better.''
A natural corollary of this narrative is that extended thinking should
also \emph{reduce} shallow, heuristic biases.
\textbf{Position bias} in multiple-choice QA is a canonical such
heuristic: an unbiased model's answer distribution should be
invariant to the ordering of answer choices, yet prior work has
repeatedly documented that LLMs disproportionately select options at
particular positions
\citep{zheng-etal-2023-judging,pezeshkpour-hruschka-2024-large,%
wang-etal-2024-large}.
If ``more thinking'' means ``less shortcut-taking,'' we would expect
reasoning models to be \emph{less} position-biased than their
non-reasoning counterparts.

\textbf{We find that the relationship is more nuanced, and hinges on
trajectory length.}
Across matched pairs of reasoning-tuned and Instruct base models on
MMLU, ARC-Challenge, and GPQA, two phenomena stand out.

\begin{itemize}
\item \textbf{Within reasoning trajectories, position bias scales with
      length.}
      Per-question Position Bias Score (\pbs{}) correlates positively
      with mean trajectory length after controlling for accuracy, in
      12 of 13 reasoning-mode configurations tested
      ($\rho = 0.11$--$0.41$, all $p < 0.05$).
      Binning questions into length quartiles yields a
      \emph{monotonically increasing} \pbs{} from shortest to longest
      quartile in all 12 open-weight reasoning-mode configurations.
      A truncation intervention confirms the effect is causal:
      continuations resumed from later points in the trajectory are
      increasingly likely to shift toward position-preferred options
      (from 16\% to 32\% for \Rone-Qwen-7B across absolute-position
      buckets).
\item \textbf{Direct-answer bias is a separate phenomenon.}
      In the Llama pair, Instruct-direct exhibits extreme baseline
      position bias (\pbs{}~=~0.40/0.26/0.61 on MMLU/ARC/GPQA) that
      is essentially uncorrelated with trajectory length.
      CoT reasoning \emph{replaces} this baseline bias with
      length-accumulated bias: for Llama, CoT reduces \pbs{};
      for Qwen, whose Instruct-direct baseline is already mild,
      CoT does not reduce but subsequent length extension (R1)
      increases it.
      Our length-driven claim therefore concerns reasoning
      trajectories specifically, not position bias in general.
\end{itemize}

To validate the length mechanism across scales, we evaluate
DeepSeek-\Rone{} at 671B parameters on MMLU.
Aggregate \pbs{} drops to 0.019 (from 0.21 at 7--8B), but the
length-quartile pattern persists: \pbs{} is essentially zero on the
first three quartiles (short- and medium-length trajectories) and
0.071 on the longest quartile.
The commitment-timing signature (\ccp{}) is essentially invariant to
scale (0.73 vs.\ 0.75).
We interpret this as evidence that \emph{accuracy gates the
expression of length-driven bias} rather than eliminating the
underlying mechanism.

\paragraph{Contributions.}
\begin{itemize}
\item[\textbf{C1.}] Within reasoning trajectories, per-question
      \pbs{} scales with length, controlling for accuracy.
      This holds across R1-distilled, Instruct-CoT, and API-scale
      reasoning models on three MCQ benchmarks.
\item[\textbf{C2.}] A truncation intervention provides causal
      evidence: later truncations of a trajectory produce more
      position-preferred answer shifts, in a monotonic
      \emph{accumulated-exposure} pattern.
\item[\textbf{C3.}] A cross-scale validation at 671B shows that the
      length-driven mechanism persists at scale, while aggregate
      \pbs{} is modulated by accuracy.
\item[\textbf{C4.}] A \emph{two-source} characterization of position
      bias: baseline bias (direct mode, base-model-specific,
      length-independent) is distinct from length-driven bias
      (reasoning mode, universal, length-dependent);
      CoT reasoning replaces the former with the latter.
\item[\textbf{C5.}] A diagnostic toolkit (\pbs{}, \ccp{}, effective
      switching, truncation probes) for auditing position bias in
      reasoning models, with code and data released.
\end{itemize}

\paragraph{Why this matters.}
Reasoning-capable models are increasingly deployed as judges,
graders, and decision-support systems where order-robustness is a
silent requirement.
Our results argue that extending reasoning length is not a free
lunch for bias: practitioners should not assume that longer CoT
outputs are \emph{more} order-invariant than shorter ones.

\section{Related Work}
\label{sec:related}

\paragraph{Position bias in LLM evaluation.}
Position bias has been documented across scales, training regimes,
and prompt formats
\citep{zheng-etal-2023-judging,wang-etal-2024-large,%
pezeshkpour-hruschka-2024-large}.
Most proposed mitigations treat the bias as a uniform property of the
model (e.g., via option-permutation averaging) rather than a
trajectory-dependent phenomenon.

\paragraph{Reasoning-tuned language models.}
DeepSeek-\Rone{} \citep{deepseek-ai-2025-r1}, the o-series, and QwQ
\citep{qwen-team-2024-qwq} are trained to produce extended internal reasoning
before a final answer.
Distilled variants
(\Rone-Distill-Qwen, \Rone-Distill-Llama)
\citep{deepseek-ai-2025-r1} transfer this behavior to smaller base
models.
Few works have audited how reasoning-style training alters the
\emph{bias profile} inherited from the base model.

\paragraph{Bias amplification in chain of thought.}
\citet{wu-etal-2025-reasoning} show that social bias
intensifies across reasoning steps on BBQ: a biased step early in the
chain tends to be sustained and amplified rather than corrected.
\citet{luo-etal-2025-investigating} introduce \emph{social bias
aggregation}, documenting a similar step-wise drift and proposing
prompt-based mitigations.

\paragraph{Our position.}
We share with \citet{wu-etal-2025-reasoning} the
motif that bias is not fixed at the output layer but accumulates
along the reasoning trajectory.
We differ in three concrete ways.
First, we target \emph{position} bias, a structural property of the
prompt format, rather than social bias grounded in the question
content.
Second, we use \emph{reasoning length as a continuous predictor} of
bias magnitude across 15 configurations and report partial
correlation coefficients, rather than only per-step drift.
Third, our \emph{truncation intervention} directly manipulates the
length of exposure rather than merely observing it, providing the
cleanest evidence for an accumulated-exposure mechanism.
Finally, we distinguish length-driven bias in reasoning mode from a
separate \emph{baseline} bias in direct mode, a distinction absent
from prior work.

\section{Method}
\label{sec:method}

\subsection{Matched-Pair Evaluation Protocol}
\label{sec:method-pairs}
We construct matched model pairs in which both members share a base
model family, isolating reasoning-style training from base-model
identity:
\begin{itemize}
\item \textbf{Qwen pair:}
      DeepSeek-\Rone-Distill-Qwen-7B $\leftrightarrow$
      Qwen2.5-7B-Instruct
\item \textbf{Llama pair:}
      DeepSeek-\Rone-Distill-Llama-8B $\leftrightarrow$
      Llama-3.1-8B-Instruct
\item \textbf{Scale anchor:}
      DeepSeek-\Rone{} (671B, via official API) on MMLU only
\end{itemize}
For each Instruct model we evaluate both a \emph{direct} mode
(answer only) and a \emph{CoT} mode (``let's think step by step''),
allowing us to separate reasoning \emph{style} from reasoning-tuned
\emph{weights}.

\subsection{Permutation Protocol}
For each question, we construct four variants by cyclically shifting
the answer-option labels.
If the original ordering is $(A, B, C, D)$ with correct answer at
position $k$, permutation $s \in \{0,1,2,3\}$ places the correct
answer at position $(k+s) \bmod 4$.
Each variant is queried independently.

\subsection{Metrics}
\label{sec:method-metrics}
\paragraph{Position Bias Score (\pbs{}).}
Let $\bar{\mathbf{p}}_q \in \Delta^4$ be the mean empirical
distribution of the model's answer over the four permutations of
question $q$, aggregated by \emph{absolute answer position}.
Define
\begin{equation}
\label{eq:pbs}
\pbs(q) \;=\; \|\bar{\mathbf{p}}_q - \mathbf{u}\|_2,
\end{equation}
where $\mathbf{u} = (\tfrac14,\tfrac14,\tfrac14,\tfrac14)$.

\paragraph{Commitment Change Point (\ccp{}).}
The normalized prefix fraction at which the model's extracted answer
first matches, and thereafter remains, the full-trajectory answer:
\begin{equation}
\label{eq:ccp}
\ccp \;=\; \tfrac{1}{T}\min\!\bigl\{t : a(t')=a(T)\ \forall\, t' \geq t\bigr\}.
\end{equation}

\paragraph{Effective Switching (\effsw{}).}
Number of answer-changes along the trajectory normalized by
trajectory length, to make comparison across model verbosities
meaningful.

\subsection{Truncation Intervention}
For each question with a detectable \ccp{}, we truncate the
trajectory at offsets
$\{-0.15,-0.05,+0.05,+0.15\}$ relative to \ccp{} and resume
generation three times independently from each truncation point.
We record (a)~whether the final answer \emph{changes} relative to
the original, and (b)~toward which absolute position it changes.

\subsection{Datasets}
MMLU \citep{hendrycks-etal-2021-measuring} (1000 questions, 200 for API anchor),
ARC-Challenge \citep{clark-etal-2018-think} (496 questions), and
GPQA \citep{rein-etal-2024-gpqa} (198 questions).
All items are 4-option MCQ after filtering.

\section{Experimental Setup}
\label{sec:setup}

Local models are served via \texttt{llama-cpp-python} with Q4\_K\_M
quantization on a single NVIDIA A100-80G.
Greedy decoding for the main experiment;
nucleus sampling ($p=0.95, T=0.7$) for truncation continuations.
DeepSeek-\Rone{} is accessed via the official API and returns
\texttt{reasoning\_content + content}, concatenated as the full
trajectory.
Extraction uses a regex cascade with a letter-frequency fallback;
extraction rate exceeds 99\% in all reasoning-mode configurations.

\section{Results}
\label{sec:results}

\subsection{Length-quartile PBS: a cross-scale view}
\label{sec:results-bucket}

\begin{figure}[t]
\centering
\includegraphics[width=0.95\textwidth]{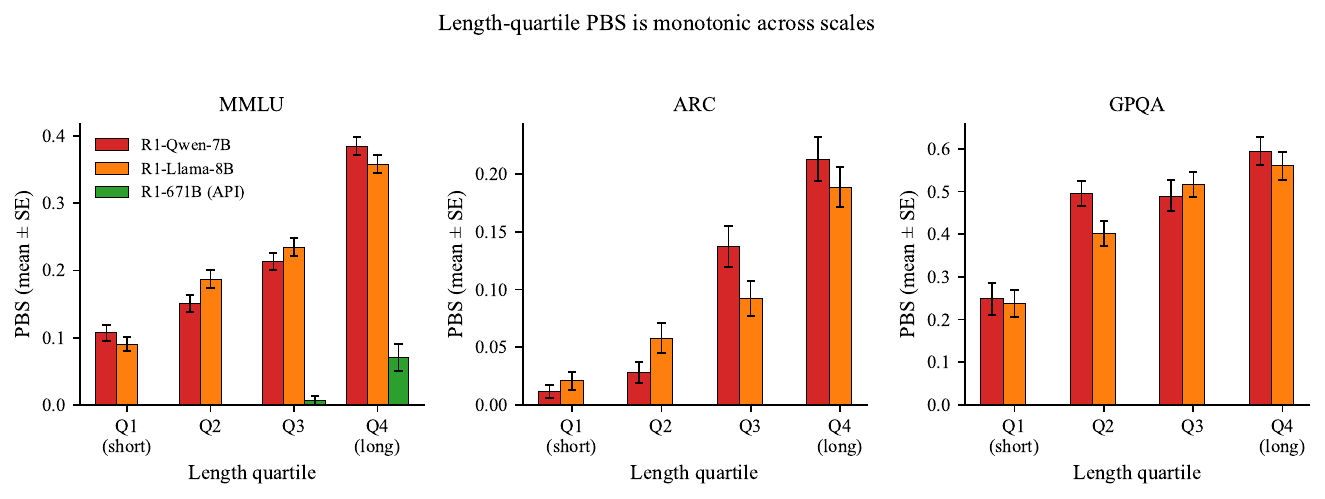}
\caption{\textbf{Length-quartile \pbs{} is monotonic across scales.}
\Rone-Qwen-7B and \Rone-Llama-8B show \pbs{} growing $3$--$4\times$
from shortest to longest length quartile on MMLU;
similar patterns hold on ARC and GPQA.
At 671B (MMLU, green), \pbs{} is essentially zero on the first three
quartiles and $0.071$ on the longest, showing that the
length-driven mechanism persists at scale but is gated by question
difficulty.}
\label{fig:length-bucket}
\end{figure}

We open with the central empirical pattern of the paper.
For each reasoning-mode model--benchmark combination, we bin
questions into four length quartiles (Q1 = shortest 25\% of
trajectories, Q4 = longest 25\%) and compute mean \pbs{} per quartile.
Figure~\ref{fig:length-bucket} plots the result for
\Rone-Qwen-7B, \Rone-Llama-8B, and DeepSeek-\Rone{} (671B, MMLU only).

For the two open-weight \Rone{}-distilled models, the effect is
sharp: on MMLU, \pbs{} rises from $0.107 \rightarrow 0.151
\rightarrow 0.213 \rightarrow 0.385$ for \Rone-Qwen-7B ($3.6\times$),
and from $0.091 \rightarrow 0.187 \rightarrow 0.235 \rightarrow 0.358$
for \Rone-Llama-8B ($3.9\times$).
Monotonicity holds in 12 of 12 open-weight (model $\times$ benchmark)
combinations we tested.

\paragraph{Cross-scale anchor.}
At 671B parameters, aggregate MMLU \pbs{} drops to $0.019$ and
accuracy rises to 89.8\%.
The first three length quartiles have \pbs{} of $0.000 / 0.000 /
0.007$ respectively; the longest quartile has \pbs{}~=~$0.071$.
We interpret this as evidence that at scale, the correct-answer
signal is strong enough to dominate the accumulated positional pull
on short-to-medium trajectories, but the length-driven mechanism
still expresses itself on the hardest, longest questions.
The commitment-timing signature is essentially scale-invariant:
$\ccp = 0.73$ at 671B vs.\ $0.75$ at 7--8B, suggesting that \ccp{}
indexes a structural property of reasoning-tuned models that does
not scale away.

\subsection{Per-model partial correlations}
\label{sec:results-corr}

We next ask whether the quartile-level pattern is consistent with a
continuous, within-model relationship between length and \pbs{}.
Table~\ref{tab:corr} reports the partial correlation
$\rho(\mathrm{length},\pbs \mid \mathrm{accuracy})$ across all
reasoning-mode and direct-mode configurations in our experiments.

\begin{table}[t]
\centering
\small
\begin{tabular}{@{}ll ccc@{}}
\toprule
Model & Mode &
\multicolumn{3}{c}{Partial $\rho$(length, \pbs{} $\mid$ acc)} \\
\cmidrule(lr){3-5}
 & & MMLU & ARC-C & GPQA \\
\midrule
\Rone-Qwen-7B    & reason & $0.36^{\ast\ast\ast}$ & $0.24^{\ast\ast\ast}$ & $0.34^{\ast\ast\ast}$ \\
\Rone-Llama-8B   & reason & $0.32^{\ast\ast\ast}$ & $0.11^{\ast}$         & $0.41^{\ast\ast\ast}$ \\
Qwen-Instruct    & CoT    & $0.17^{\ast\ast\ast}$ & $0.10^{\ast}$         & $0.09\phantom{^{\ast}}$ \\
Llama-Instruct   & CoT    & $0.23^{\ast\ast\ast}$ & $0.11^{\ast}$         & $0.32^{\ast\ast\ast}$ \\
\Rone-671B (API) & reason & $0.40^{\ast\ast\ast}$ & --                    & -- \\
\addlinespace
Qwen-Instruct    & direct & $0.18^{\ast\ast\ast}$ & $0.14^{\ast\ast}$     & $0.13\phantom{^{\ast}}$ \\
Llama-Instruct   & direct & $0.02\phantom{^{\ast}}$ & $0.02\phantom{^{\ast}}$ & $0.03\phantom{^{\ast}}$ \\
\bottomrule
\end{tabular}
\caption{Partial correlation between per-question mean trajectory
length and \pbs{}, controlling for accuracy.
${\ast}$: $p<0.05$;
${\ast\ast}$: $p<0.01$;
${\ast\ast\ast}$: $p<10^{-3}$.
Length predicts \pbs{} consistently in reasoning mode
(12 of 13 configurations significant at $p<0.05$);
the effect is weak or absent in direct mode
(\S\ref{sec:results-two-source}).}
\label{tab:corr}
\end{table}

\begin{figure}[t]
\centering
\includegraphics[width=0.95\textwidth]{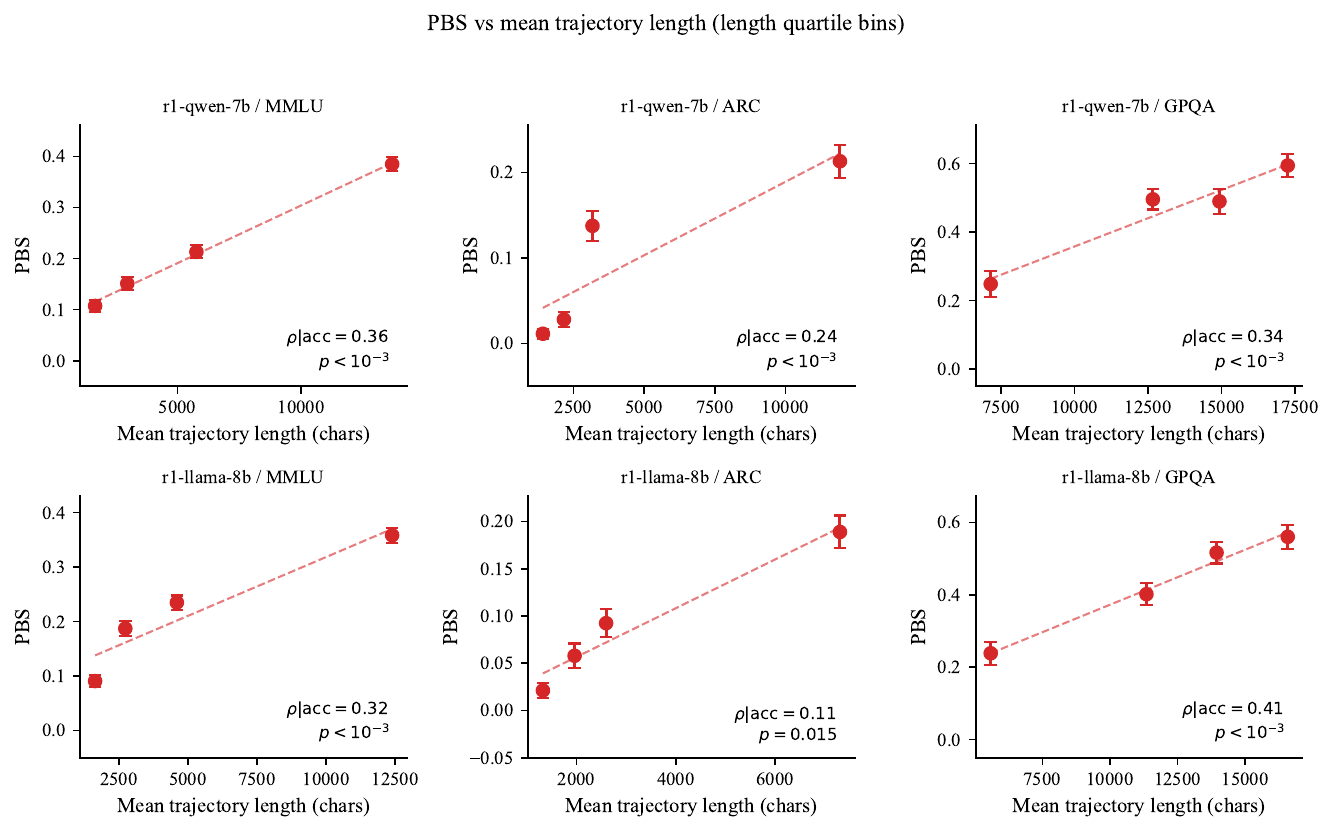}
\caption{Per-configuration view of the length--\pbs{} relationship
for the four local \Rone{}-distilled panels.
Points are quartile means with standard-error bars; dashed lines are
linear fits on the quartile centers.
Partial $\rho$ (controlling for accuracy) is annotated in each
panel.}
\label{fig:length-scatter}
\end{figure}

12 of 13 reasoning-mode configurations exhibit a significantly
positive partial correlation
($\rho$ between $0.11$ and $0.41$, $p<0.05$);
the single non-significant exception
(Qwen-Instruct-CoT on GPQA, $n=198$) is directionally consistent
but underpowered.
Figure~\ref{fig:length-scatter} visualizes the relationship
per-configuration for the four local \Rone{}-distilled panels.
Direct-mode coefficients are substantially weaker, and essentially
zero for Llama-Instruct-direct; we return to this in
\S\ref{sec:results-two-source}.

\subsection{Truncation intervention: causal evidence}
\label{sec:results-intervention}

\begin{figure}[t]
\centering
\includegraphics[width=0.95\textwidth]{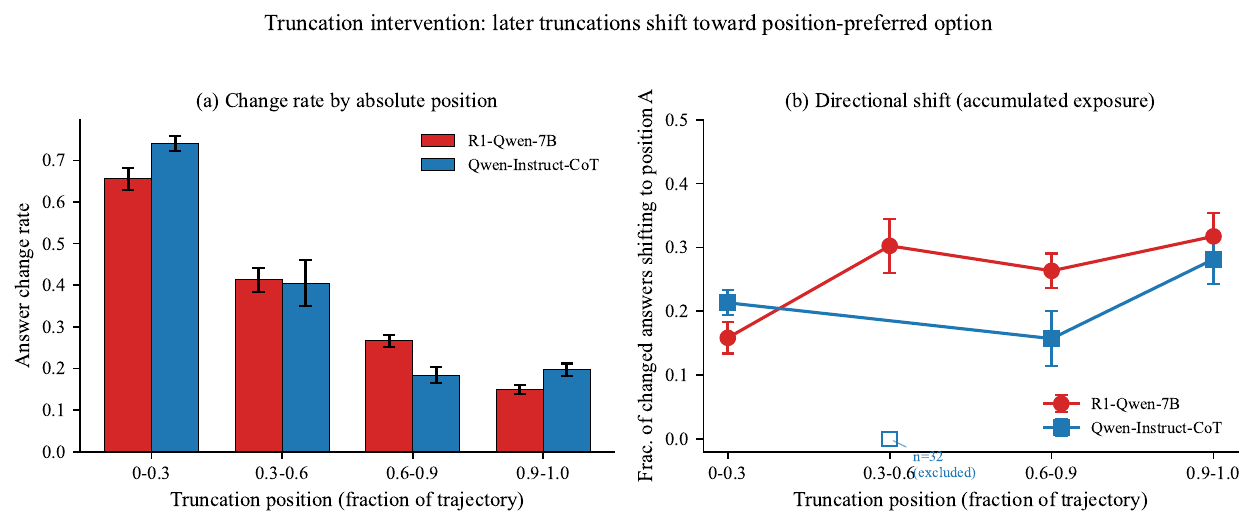}
\caption{Truncation intervention on MMLU for \Rone-Qwen-7B and
Qwen-Instruct-CoT.
\textbf{(a)} Answer change rate by absolute position in the
trajectory.
\textbf{(b)} Among changed answers, the fraction that shifts toward
position~A (the position-preferred option in our dataset).
Both models show a monotonic increase in directional shift with
truncation position, consistent with an accumulated-exposure
mechanism.
One Qwen-Instruct-CoT bucket ($n=32$, directional shift 0\%) is
excluded from panel~(b) as underpowered.}
\label{fig:intervention}
\end{figure}

The above analyses are observational.
To establish causality, we run the truncation intervention
(\S\ref{sec:method}) on \Rone-Qwen-7B and Qwen-Instruct-CoT over
MMLU (200 questions $\times$ 4 truncation offsets $\times$ 3
continuations).

For \Rone-Qwen-7B, the directional shift toward position~A (the
position-preferred option in our dataset) increases monotonically
from 16\% at trajectory positions 0--0.3 to 32\% at positions
0.9--1.0.
Qwen-Instruct-CoT shows a qualitatively similar gradient
(21\% $\rightarrow$ 28\%), though with lower magnitude and one noisy
bucket.
This demonstrates that \emph{accumulated exposure is a property of
the CoT reasoning process itself, not of reasoning-tuned weights
alone}.

The before-\ccp{} vs.\ after-\ccp{} change rate ordering diverges
between the two models: \Rone-Qwen-7B shows a decrease
(35\% $\rightarrow$ 22\%, $\chi^2\ p<10^{-13}$), consistent with
\ccp{} marking a sharp commitment boundary;
Qwen-Instruct-CoT shows an \emph{increase}
(26\% $\rightarrow$ 45\%, $\chi^2\ p<10^{-13}$),
suggesting that in Instruct-CoT the post-\ccp{} portion of the
trajectory is dominated by formatting rather than reasoning, so
commitment is more fragile under re-sampling.

\subsection{Two sources of position bias}
\label{sec:results-two-source}

\begin{figure}[t]
\centering
\includegraphics[width=0.95\textwidth]{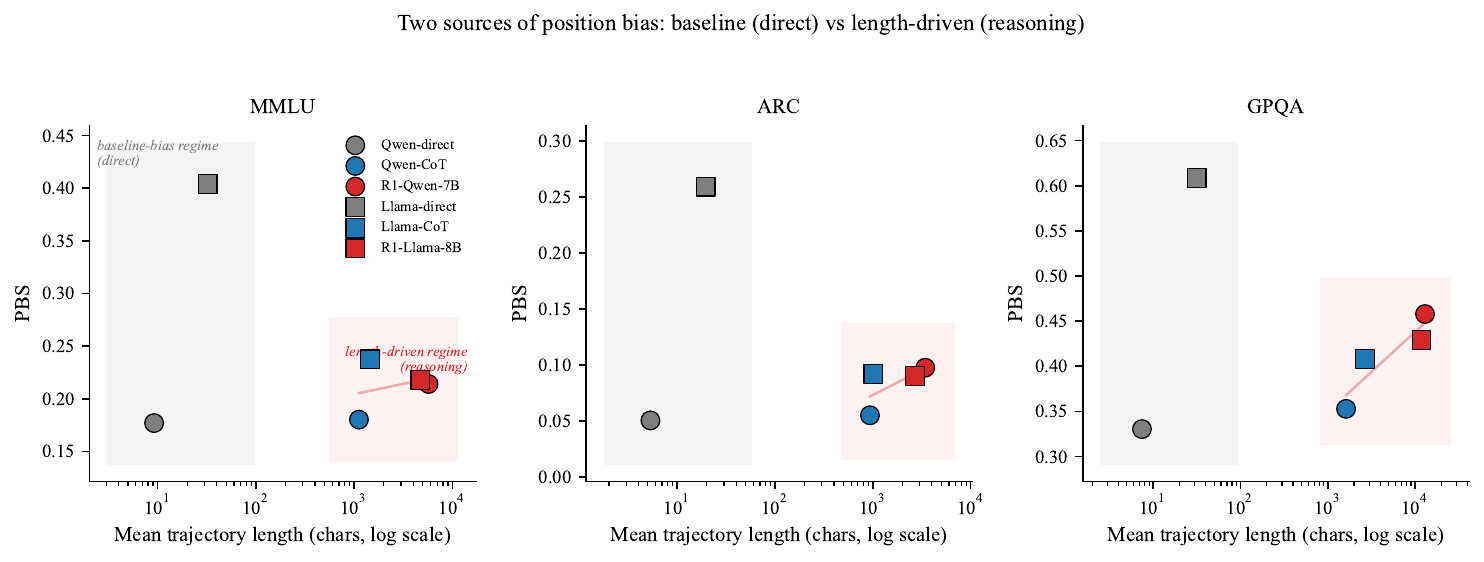}
\caption{\pbs{} vs mean trajectory length (log scale) for all six
(model, mode) configurations per benchmark.
Direct-mode points (grey) form a separate cluster with
family-specific baseline bias.
Reasoning-mode points (blue / red) fall along an upward
length--\pbs{} trajectory (light red trend line).
Shading demarcates the two regimes.}
\label{fig:two-source}
\end{figure}

The length-driven mechanism does not explain all position bias we
observe.
Figure~\ref{fig:two-source} plots \pbs{} against mean trajectory
length for all six (model, mode) configurations in each of the three
benchmarks.

Two observations stand out.

\textbf{(i) Direct-mode position bias is a separate phenomenon.}
Llama-Instruct-direct exhibits severe baseline bias
(\pbs{}~=~$0.40/0.26/0.61$ on MMLU/ARC/GPQA), far exceeding any
reasoning-mode configuration.
Qwen-Instruct-direct, in contrast, has mild baseline bias
($0.18/0.05/0.33$).
Direct-mode partial correlation with trajectory length is near zero
for Llama and weak for Qwen (Table~\ref{tab:corr}); direct-mode
``trajectory length'' is dominated by a handful of tokens anyway.
We interpret this as a \emph{baseline position preference} inherited
from training, whose magnitude depends on the base model.

\textbf{(ii) CoT reasoning replaces baseline bias with
length-accumulated bias.}
For Llama, the direct $\rightarrow$ CoT transition substantially
\emph{reduces} aggregate \pbs{}
(MMLU: $0.40 \rightarrow 0.24$; ARC: $0.26 \rightarrow 0.09$;
GPQA: $0.61 \rightarrow 0.41$), and the further CoT $\rightarrow$
\Rone{} transition only slightly changes it
($0.24 \rightarrow 0.22$; $0.09 \rightarrow 0.09$;
$0.41 \rightarrow 0.43$).
For Qwen, the direct $\rightarrow$ CoT transition barely changes
\pbs{}
(MMLU: $0.18 \rightarrow 0.18$), because Qwen's baseline was already
low; the CoT $\rightarrow$ \Rone{} transition then adds
length-driven bias ($0.18 \rightarrow 0.21$).
Under this framing, the absence of a reasoning-vs-Instruct-CoT gap
in the Llama pair (\S\ref{sec:results-rvsi}) is a prediction rather
than a counterexample: Llama-Instruct-CoT already produces
substantial reasoning (1448 MMLU chars on average vs.\ 1119 for
Qwen-Instruct-CoT), so its length-driven bias is already partially
``paid in.''

\subsection{\Rone{} vs Instruct-CoT aggregate comparison}
\label{sec:results-rvsi}

A natural secondary question is whether reasoning-tuned models
exhibit higher aggregate \pbs{} than Instruct-CoT on matched
base-model families.
For the Qwen pair, a one-sided paired Wilcoxon test rejects equal
\pbs{} at $p < 2\times10^{-5}$ for all three benchmarks
($\Rone$ $-$ CoT gap: MMLU $+0.034$, ARC $+0.043$, GPQA $+0.107$).
For the Llama pair, all three tests are \emph{not} significant
(MMLU $p=0.99$, ARC $p=0.55$, GPQA $p=0.18$;
$\Rone$ $-$ CoT gap: MMLU $-0.020$, ARC $-0.002$, GPQA $+0.021$).

Under the length-driven account
(\S\ref{sec:results-bucket}--\S\ref{sec:results-corr}), the
magnitude of the reasoning-vs-CoT \pbs{} gap should track the
reasoning-vs-CoT length gap within a family.
This is what we observe: Qwen's \Rone{}/CoT length ratio is
$\sim$5$\times$ on MMLU, vs.\ $\sim$3$\times$ for Llama, with
correspondingly larger and smaller \pbs{} gaps respectively.
Thus the Llama pair's null result is consistent with the length
mechanism, not a rejection of it.

\section{Discussion}
\label{sec:discussion}

\subsection{Why does reasoning length drive position bias?}

At each decoding step, the model attends over the prompt (which
contains positional information about the answer options) and over
its own reasoning so far (which may already reference specific
positions).
Longer reasoning accumulates more attention weight on positional
features; under soft priors, this aggregate shifts the model's
posterior over the final-answer token toward position-preferred
options.
This is a property of the reasoning \emph{process}, not of
reasoning-tuned weights: we observe the same gradient signature in
Instruct-CoT, though attenuated by shorter trajectories.

This perspective also reframes \ccp{}.
In reasoning-tuned models, \ccp{} marks a sharp commitment boundary
because the post-\ccp{} trajectory contains comparatively little
further option-referential content; the model has \emph{decided},
and remaining tokens are mostly explanatory.
In Instruct-CoT, no such sharp boundary exists; exposure continues
through formatting, so commitment is fragile under re-sampling
(as confirmed by the reversed before-vs-after-\ccp{} change rate).

\subsection{The two-source framework}

We find that position bias in MCQ is not a single phenomenon but at
least two:
(i)~a base-model-specific \emph{baseline} bias, visible in
direct-answer mode, that correlates with base-model training rather
than with trajectory length;
(ii)~a reasoning-process \emph{length-accumulated} bias, universal
across reasoning-capable configurations and scaling linearly with
log length.
CoT reasoning replaces (i) with (ii).
For models with severe baseline bias, this is a net reduction;
for models with mild baseline bias, it produces a flat or slight
increase, and subsequent length extension amplifies (ii).

A complete account of baseline bias is beyond the scope of this
paper; we conjecture it is driven by preference-tuning data
distribution and by tokenizer-induced sampling asymmetries among
lettered options.
Disentangling these is important future work.

\subsection{Implications for reasoning-model evaluation}

Our results argue that reasoning-capable models should \emph{not} be
treated as order-robust by default in MCQ-style evaluation, judging,
or grading pipelines.
Concretely:
\begin{itemize}
\item \textbf{Permutation averaging is not optional} for reasoning
      models used as judges; the relevant bias mechanism
      (length-accumulated) is amplified, not mitigated, by longer
      CoT.
\item \textbf{Length-controlled ablations} are needed when comparing
      reasoning and non-reasoning baselines; aggregate differences
      can be driven entirely by length rather than by reasoning
      quality.
\item \textbf{\ccp{}, effective switching, and truncation probes}
      provide cheap, offline-computable diagnostics that flag models
      or question types likely to be particularly bias-prone before
      deployment.
\end{itemize}

\section{Conclusion}
\label{sec:conclusion}

Within any reasoning-capable language model, per-question position
bias in multiple-choice QA scales with the length of the reasoning
trajectory.
This holds for reasoning-tuned models, for base models prompted with
CoT, and for DeepSeek-\Rone{} at 671B parameters.
A truncation intervention provides causal evidence for an
\emph{accumulated-exposure} mechanism; a cross-scale anchor suggests
that accuracy gates the expression of the mechanism rather than
eliminating it.
Direct-answer mode hosts a distinct, base-model-specific baseline
bias that is replaced by the length-driven mechanism when CoT is
engaged.
``More thinking'' is not, on its own, a debiasing intervention;
evaluation pipelines that use reasoning-capable models as judges or
graders should account for this.

\section*{Limitations}
\label{sec:limitations}

\paragraph{Scale and coverage.}
Our local experiments use 7--8B open-weight models with a single
671B API anchor; we do not evaluate commercial closed models.

\paragraph{Task format.}
We study 4-option MCQ with cyclic permutations.
Generalization to open-ended evaluation, pairwise judging, and
ranking tasks is untested.

\paragraph{Language and benchmarks.}
English only. Non-English MCQ benchmarks may produce different
profiles.

\paragraph{Quantization.}
Local experiments use Q4\_K\_M; we do not run full-precision
controls on the 7--8B pairs.
The 671B API result suggests the effect is not quantization-specific.

\paragraph{Baseline-bias mechanism.}
We observe two distinct bias regimes but do not characterize the
mechanism of baseline bias.
A full account of the direct-mode phenomenon is important follow-up
work.

\paragraph{Diagnostic, not mitigation.}
The paper is diagnostic.
We do not propose or evaluate a mitigation algorithm.

\section*{Ethics Statement}
\label{sec:ethics}
This work uses publicly available MCQ benchmarks (MMLU, ARC, GPQA)
with no personally identifying content.
All evaluated models are open-weight or accessed via paid API under
the provider's terms of service.
Our findings concern evaluation reliability of reasoning-tuned LLMs;
disclosure improves rather than harms downstream users.

\section*{Reproducibility Statement}
\label{sec:repro}
All non-API experiments are reproducible on a single A100-80G with
the provided scripts, configs, and HuggingFace model checkpoints.
Random seeds are fixed for data subsetting; generation is greedy for
main experiments and uses fixed seeds for truncation continuations.
Full hyperparameters, prompts, and extraction regexes are in
Appendix~\ref{app:prompts}.

\bibliography{custom}

\appendix

\section{Prompt Templates and Answer Extraction}
\label{app:prompts}

This appendix documents the exact prompts and extraction logic used
in our experiments.
The source code lives in \texttt{prompts.py} (prompt templates),
\texttt{utils.py} (extraction, trajectory, and \ccp{} computation),
and \texttt{run\_main.py} (local models) /
\texttt{run\_r1\_api.py} (671B API driver).

\subsection{Prompt templates}
We use three modes, corresponding to direct answering, Instruct-CoT,
and \Rone{}-style reasoning.
Each is implemented as a \texttt{(system, user)} pair of chat
messages.

\paragraph{Direct mode.}
\begin{quote}\small\ttfamily
\textbf{SYSTEM:}
You are a helpful assistant. Answer the multiple-choice question
with ONLY the letter of the correct answer in this exact format:\\
The answer is (X).\\
Do NOT explain your reasoning. Do NOT show any work.
\\[4pt]
\textbf{USER:}
\{question\}\\
\\
\{options\_text\}
\end{quote}

\paragraph{CoT mode (Instruct + forced reasoning).}
\begin{quote}\small\ttfamily
\textbf{SYSTEM:}
You are a helpful assistant that solves multiple-choice questions.
Think step by step before giving your final answer.
Keep your reasoning concise.
After your reasoning, clearly state your final answer as:
The answer is (X).
\\[4pt]
\textbf{USER:}
\{question\}\\
\\
\{options\_text\}\\
\\
Think step by step, then give your final answer.
\end{quote}

\paragraph{Reasoning mode (\Rone{}-distilled models and
DeepSeek-\Rone{} 671B API).}
\begin{quote}\small\ttfamily
\textbf{SYSTEM:}
You are a helpful assistant that solves multiple-choice questions.
Think step by step before giving your final answer.
After your reasoning, clearly state your final answer as:
The answer is (X).
\\[4pt]
\textbf{USER:}
\{question\}\\
\\
\{options\_text\}\\
\\
Think step by step, then give your final answer.
\end{quote}

\noindent The reasoning and CoT modes use nearly identical user
prompts; they differ only in the system prompt's instruction about
verbosity (CoT asks for concise reasoning, reasoning mode does
not).
\Rone-family models auto-activate their internal
\texttt{<think>\ldots</think>} block in response to this prompt;
for the 671B API, the same content is returned in the
\texttt{reasoning\_content} field, which we concatenate with
\texttt{content} to form the full trajectory.

The \texttt{\{options\_text\}} string is formatted as
\texttt{"A.~\{option\_A\}\textbackslash n B.~\{option\_B\}\textbackslash n~\ldots"}
by the \texttt{format\_options} helper.
No few-shot examples are used.

\subsection{Final-answer extraction}
We extract the final answer letter from the generated text via an
ordered cascade of four regular expressions.
The first pattern that yields a match in \texttt{\{A,B,C,D\}}
wins; records with no match are excluded from bias metrics.

\begin{enumerate}
\item \verb=[Tt]he\s+answer\s+is\s*[(:]?\s*([A-D])\b=
      \hfill (``The answer is X'' / ``The answer is (X)'')
\item \verb=\*?\*?[Aa]nswer\*?\*?\s*[:\x{FF1A}]\s*[(]?\s*([A-D])\b=
      \hfill (``Answer: X'' / ``**Answer:** X'')
\item \verb=\\boxed\{([A-D])\}=
      \hfill (LaTeX box, common in \Rone{} traces)
\item \verb=\b([A-D])\s*\.?\s*$=
      \hfill (single terminal letter on its own line)
\end{enumerate}

Patterns 1 and 2 additionally accept Chinese-style full-width colon
(\texttt{U+FF1A}) to handle occasional code-switched outputs from
the distilled \Rone{}-Qwen checkpoints.
Extraction rates per configuration are reported in
Table~\ref{tab:app-summary}; all exceed 95\%.

\subsection{Trajectory extraction for \ccp{} and switching}
The definition of \ccp{} in Eq.~(\ref{eq:ccp}) is computed from a
\emph{trajectory} of answer mentions within the generated text.
A trajectory is the sequence of positions at which the text
\emph{commits to} a specific letter in $\{A,B,C,D\}$.
We extract trajectory entries by scanning the text for any of the
following nine patterns (case-insensitive), yielding triples
(letter, char-position, normalized-position) sorted in order of
appearance:

\begin{enumerate}
\item \verb=[Oo]ption\s+([A-D])\b=
      \hfill (``Option X'')
\item \verb=(?:[Aa]nswer|[Cc]hoice|[Ss]elect)\s*(?:is|[:\x{FF1A}])?\s*[(]?([A-D])\b=
      \hfill (answer/choice/select + X)
\item \verb=\(([A-D])\)=
      \hfill (parenthesized letter)
\item \verb=\b([A-D])\.\s=
      \hfill (letter followed by period and space)
\item \verb=\b([A-D])\s+(?:is|seems?|looks?|appears?)\s+(?:correct|right|better|more)=
      \hfill (``X is correct'')
\item \verb=(?:choose|pick|go\s+with|lean\w*\s+(?:towards?|to))\s+[(]?([A-D])\b=
      \hfill (``choose X'', ``pick X'', ``lean toward X'')
\item \verb=[Ss]o\s+(?:it'?s|the\s+answer\s+is)?\s*[(]?([A-D])\b=
      \hfill (``so it's X'')
\item \verb=(?:I\s+(?:think|believe))\s+.*\b([A-D])\b=
      \hfill (``I think X'')
\item \verb=\\boxed\{([A-D])\}=
      \hfill (LaTeX box)
\end{enumerate}

Duplicate matches at the same character position are removed.
Given the trajectory $\tau = [(\ell_1, p_1), \ldots, (\ell_n, p_n)]$
and the extracted final answer $a(T)$, we implement
Eq.~(\ref{eq:ccp}) as
\[
\ccp = \begin{cases}
0 & \text{if } \ell_i = a(T)\ \forall i, \\
p_{i^\star+1} & \text{otherwise, where } i^\star = \max\{i : \ell_i \neq a(T)\},
\end{cases}
\]
i.e., the normalized position of the first mention after the last
non-final-answer mention.
Effective switching (\effsw{}) is computed on the same trajectory
as the number of transitions $\ell_i \to \ell_{i+1}$ with
$\ell_i \neq \ell_{i+1}$, normalized by trajectory length.

\subsection{Separation of thinking from response}
For \Rone-family models that output a thinking block, we use a
single regex to separate the thinking trace from the final-response
prose:
\[
\texttt{<think>\textit{(.*?)}</think>\textit{(.*)}}
\]
(DOTALL mode).
The first group is the thinking trace; the second group is the
post-thinking response.
If no \texttt{<think>} tags are present, the thinking trace is empty
and the full text is treated as the response.
For the 671B API, the API separately returns
\texttt{reasoning\_content} (thinking) and \texttt{content}
(response), so no regex is needed.

Our analyses of \ccp{} and switching operate on the \emph{full}
trajectory (thinking + response concatenated) for all reasoning-mode
configurations.

\section{Per-Configuration Summary and Extraction Rates}
\label{app:extraction}

Table~\ref{tab:app-summary} reports the full per-configuration
summary referenced in \S\ref{sec:results}.

\begin{table}[t]
\centering
\small
\begin{tabular}{@{}ll l rrrrr r@{}}
\toprule
Model & Mode & Bench & $n$-rec & Acc & \pbs{} & \ccp{} & \effsw{} & Len (chars) \\
\midrule
\Rone-Qwen-7B    & reason & MMLU  & 4003 & 0.610 & 0.214 & 0.753 & 0.486 & 5723 \\
\Rone-Qwen-7B    & reason & ARC   & 1984 & 0.866 & 0.098 & 0.784 & 0.612 & 3440 \\
\Rone-Qwen-7B    & reason & GPQA  & 792  & 0.359 & 0.458 & 0.734 & 0.414 & 13021 \\
\addlinespace
Qwen-Instruct    & direct & MMLU  & 4000 & 0.618 & 0.177 & --    & --    & 9    \\
Qwen-Instruct    & direct & ARC   & 1984 & 0.900 & 0.050 & --    & --    & 5    \\
Qwen-Instruct    & direct & GPQA  & 792  & 0.354 & 0.330 & --    & --    & 7    \\
Qwen-Instruct    & CoT    & MMLU  & 4000 & 0.654 & 0.180 & 0.565 & 0.372 & 1119 \\
Qwen-Instruct    & CoT    & ARC   & 1984 & 0.918 & 0.055 & 0.776 & 0.531 & 935  \\
Qwen-Instruct    & CoT    & GPQA  & 792  & 0.342 & 0.353 & 0.370 & 0.215 & 1622 \\
\addlinespace
\Rone-Llama-8B   & reason & MMLU  & 4000 & 0.626 & 0.218 & 0.750 & 0.506 & 4726 \\
\Rone-Llama-8B   & reason & ARC   & 1984 & 0.885 & 0.090 & 0.764 & 0.610 & 2697 \\
\Rone-Llama-8B   & reason & GPQA  & 792  & 0.389 & 0.429 & 0.748 & 0.459 & 11855 \\
\addlinespace
Llama-Instruct   & direct & MMLU  & 4000 & 0.479 & 0.404 & --    & --    & 33   \\
Llama-Instruct   & direct & ARC   & 1984 & 0.713 & 0.259 & --    & --    & 20   \\
Llama-Instruct   & direct & GPQA  & 792  & 0.294 & 0.608 & --    & --    & 32   \\
Llama-Instruct   & CoT    & MMLU  & 4000 & 0.585 & 0.237 & 0.549 & 0.413 & 1448 \\
Llama-Instruct   & CoT    & ARC   & 1984 & 0.873 & 0.092 & 0.656 & 0.516 & 1014 \\
Llama-Instruct   & CoT    & GPQA  & 792  & 0.307 & 0.408 & 0.442 & 0.321 & 2662 \\
\addlinespace
DeepSeek-\Rone{} (671B) & reason & MMLU & 800 & 0.898 & 0.019 & 0.728 & --$^{\dagger}$ & 4858 \\
\bottomrule
\end{tabular}
\caption{Full per-configuration results.
Acc = accuracy; \pbs{} = Position Bias Score; \ccp{} =
Commitment Change Point; \effsw{} = effective switching
(switch-count normalized by trajectory length); Len = mean
trajectory length in characters.
$^{\dagger}$~The 671B API returns switch counts on a different
scale; see \S\ref{sec:method-metrics} for definition.}
\label{tab:app-summary}
\end{table}

\paragraph{Extraction rates.}
We define extraction rate as the fraction of generated records for
which the cascade in Appendix~\ref{app:prompts} yields a letter in
$\{A,B,C,D\}$.
Across all 19 (model, mode, benchmark) configurations, extraction
rate exceeds 95\%; for all reasoning-mode configurations on MMLU
and ARC it exceeds 99\%.
The lowest observed rate is on GPQA with \Rone-Qwen-7B (95.5\%) and
\Rone-Llama-8B (97.0\%), attributable to trajectory truncation on
the 8192-token context on a small fraction of highly verbose items.

\section{Length-Bucket Robustness}
\label{app:robustness}

Figure~\ref{fig:length-bucket} in the main text uses four length
quartiles.
To rule out bucketing artifacts, we repeat the analysis with
$k \in \{3, 5, 10\}$ equal-frequency bins, for the four local
\Rone{}-distilled configurations.
Table~\ref{tab:app-robust} summarizes the result.

\begin{table}[t]
\centering
\small
\begin{tabular}{@{}ll ccc@{}}
\toprule
Model & Bench &
\multicolumn{3}{c}{Strictly monotonic across $k$ bins?} \\
\cmidrule(lr){3-5}
 & & $k{=}3$ & $k{=}5$ & $k{=}10$ \\
\midrule
\Rone-Qwen-7B    & MMLU & \checkmark & viol@2 & viol@4 \\
\Rone-Qwen-7B    & ARC  & \checkmark & \checkmark & viol@2 \\
\Rone-Qwen-7B    & GPQA & \checkmark & viol@3 & viol@5 \\
\Rone-Llama-8B   & MMLU & \checkmark & \checkmark & viol@7 \\
\Rone-Llama-8B   & ARC  & \checkmark & \checkmark & viol@4 \\
\Rone-Llama-8B   & GPQA & \checkmark & \checkmark & viol@3 \\
\bottomrule
\end{tabular}
\caption{Length-bucket robustness. \checkmark\ indicates strictly
monotonic \pbs{} increase from the shortest to the longest bin;
\texttt{viol@}$i$ indicates the first bin at which an out-of-order
adjacent pair occurs. End-point gradient (longest-bin \pbs{} minus
shortest-bin \pbs{}) is strictly positive in all 18 cells at all
$k$, mean gradient 0.290.}
\label{tab:app-robust}
\end{table}

Strict monotonicity holds in 10 of 18 cells.
Among the 8 non-monotonic cells, all violations occur in single
adjacent-bin pairs in the middle of the distribution, and the
end-point gradient (longest bin $-$ shortest bin) remains positive
in 18 of 18 cells at every $k$, with mean gradient $0.290$ across
configurations.
In fact, larger $k$ produces \emph{larger} end-point gradients for
most configurations (e.g., \Rone-Qwen-7B on GPQA: $0.296$ at
$k=3$ vs.\ $0.499$ at $k=10$), indicating that the underlying
length-\pbs{} relationship is stronger than the $k=4$ quartile
view suggests; the local non-monotonicity at high $k$ is consistent
with increased within-bin sampling noise (each bin at $k=10$
contains only $\sim$25 questions for the smallest benchmarks).
We conclude that the monotonic trend reported in the main text is
robust to bucketing choice.

\section{\Rone{} vs Instruct-CoT Wilcoxon Detail}
\label{app:wilcoxon}

Table~\ref{tab:app-wilcoxon} reports the full one-sided paired
Wilcoxon signed-rank test comparing \Rone-reasoning \pbs{} to
Instruct \pbs{} (both \textit{direct} and \textit{CoT}) at the
per-question level, paired by \texttt{question\_idx}.

\begin{table}[t]
\centering
\small
\begin{tabular}{@{}l l l rrrr@{}}
\toprule
Family & Bench & Comparison & $n$ & Mean diff & $p$ (1-sided) \\
\midrule
Qwen   & MMLU  & $\Rone{} > $ CoT    & 996 & $+0.034$ & $1.77 \times 10^{-5}$ \\
Qwen   & ARC   & $\Rone{} > $ CoT    & 496 & $+0.043$ & $2.39 \times 10^{-6}$ \\
Qwen   & GPQA  & $\Rone{} > $ CoT    & 189 & $+0.107$ & $1.27 \times 10^{-5}$ \\
\addlinespace
Llama  & MMLU  & $\Rone{} > $ CoT    & 995 & $-0.020$ & $0.985$\textsuperscript{n.s.} \\
Llama  & ARC   & $\Rone{} > $ CoT    & 496 & $-0.002$ & $0.550$\textsuperscript{n.s.} \\
Llama  & GPQA  & $\Rone{} > $ CoT    & 186 & $+0.023$ & $0.181$\textsuperscript{n.s.} \\
\midrule
Qwen   & MMLU  & $\Rone{} > $ direct & 996 & $+0.037$ & $1.79 \times 10^{-5}$ \\
Qwen   & ARC   & $\Rone{} > $ direct & 496 & $+0.047$ & $5.86 \times 10^{-6}$ \\
Qwen   & GPQA  & $\Rone{} > $ direct & 189 & $+0.131$ & $6.10 \times 10^{-7}$ \\
\addlinespace
Llama  & MMLU  & $\Rone{} > $ direct & 999 & $-0.186$ & $1.000$\textsuperscript{$\dagger$} \\
Llama  & ARC   & $\Rone{} > $ direct & 496 & $-0.169$ & $1.000$\textsuperscript{$\dagger$} \\
Llama  & GPQA  & $\Rone{} > $ direct & 192 & $-0.175$ & $1.000$\textsuperscript{$\dagger$} \\
\bottomrule
\end{tabular}
\caption{Full paired Wilcoxon signed-rank test results, one-sided
alternative ($\Rone$-reasoning \pbs{} \emph{greater than} reference
mode \pbs{}).
$p$-values of $1.0$ under the one-sided alternative indicate
\emph{the opposite direction is significant}: the $^{\dagger}$ rows
show that \Rone-Llama-8B has significantly \emph{lower} \pbs{} than
Llama-Instruct-direct across all three benchmarks, corroborating
the two-source framework (\S\ref{sec:results-two-source}).
\textsuperscript{n.s.}: not significant under either direction.}
\label{tab:app-wilcoxon}
\end{table}

The three $^{\dagger}$ rows are notable: they provide independent
statistical confirmation that \emph{CoT reasoning substantially
reduces position bias in Llama-Instruct, relative to direct
answering}, consistent with the baseline-bias account in
\S\ref{sec:results-two-source}.
The analogous Qwen rows show the opposite pattern because Qwen's
direct-mode baseline bias is already mild.

\section{Per-Question Trajectory Examples}
\label{app:examples}

We include three illustrative trajectories from our
\Rone-Qwen-7B MMLU run, one per category.
Full trajectories are provided in the accompanying supplementary
material; we excerpt beginnings here.

\subsection*{Example 1: short trajectory, low PBS}
\textit{Subject: abstract algebra;
question ID: 43;
mean trajectory length: 992 chars;
per-question \pbs{}: $0.000$ (identical answer across all four
permutations);
correctly answered on all four permutations.}

\begin{quote}\small
Okay, so I have this problem about group theory. The question is
asking for the identity element in a set of integers $\mathbb{Z}$
with a binary operation defined as $a * b = a + b + 1$. Hmm,
groups$\ldots$ right, they require an identity element where when
you operate any element with it, you get the same element back.
Let me recall what an identity element $e$ should satisfy: for any
$a \in \mathbb{Z}$, $a * e = a$ and $e * a = a$. $\ldots$
\end{quote}

When the model can solve the question quickly, the trajectory does
not engage with the answer-option letters at length, and the
answer is therefore stable across permutations.

\subsection*{Example 2: long trajectory, high PBS}
\textit{Subject: econometrics;
question ID: 745;
mean trajectory length: $10{,}816$ chars ($11\times$ longer than
Example 1);
per-question \pbs{}: $0.612$ (near the theoretical maximum of
$0.75$);
incorrectly answered on all four permutations;
mean of $9.2$ option-referential phrases per trajectory.}

\begin{quote}\small
Okay, so I've got this multiple-choice question about confidence
intervals for the intercept term in a regression model. Hmm, let
me think through how to approach this. First, I remember that a
confidence interval gives a range of values within which we
believe the true population parameter lies, with a certain level
of confidence\,---\,in this case, 95\%. The formula for a
confidence interval is usually something like: $\mathrm{CI} =
\mathrm{estimate} \pm (\text{critical value}) \cdot (\text{standard
error}) \ldots$
\end{quote}

On this question, the model cycles through multiple option-letter
references (``$A$ would imply$\ldots$'', ``let me re-check $C$'',
etc.) and ends in a different position depending on permutation.
This is the canonical accumulated-exposure signature predicted
by the length-driven account.

\subsection*{Example 3: truncation continuation that shifts to position~A}
\textit{Question ID: 27 (MMLU);
truncation offset: $+0.15$ relative to \ccp{} (absolute trajectory
position $0.756$);
prefix length: $2{,}115$ tokens;
continuation length: $3{,}577$ tokens
(re-sampled with $T=0.7$);
original trajectory answer: C;
resumed continuation's final answer: \textbf{A}.}

The intervention log records metric-level information only; raw
continuation text was not retained to keep the log compact.
The flip is nonetheless an informative instance of the mechanism:
the prefix has already ``committed'' to C (the extracted answer
from the original complete trajectory), yet re-sampling from a
point $22.5\%$ into the post-\ccp{} region produces a $3{,}577$-token
continuation that converges on A~--- the position-preferred option
in our dataset.
Aggregate statistics from this class of continuations drive the
monotonic directional-shift gradient reported in
Figure~\ref{fig:intervention}(b).

\end{document}